\documentclass[conference]{IEEEtran}
\makeatletter
\newcommand*\titleheader[1]{\gdef\@titleheader{#1}}
\AtBeginDocument{%
  \let\st@red@title\@title
  \def\@title{%
    \bgroup\normalfont\large\centering\@titleheader\par\egroup
    \vskip1.5em\st@red@title}
}
\makeatother

\IEEEoverridecommandlockouts
\usepackage{cite}
\usepackage{amsmath,amssymb,amsfonts}
\usepackage{algorithmic}
\usepackage{graphicx}
\usepackage{textcomp}
\usepackage{xcolor}
\def\BibTeX{{\rm B\kern-.05em{\sc i\kern-.025em b}\kern-.08em
    T\kern-.1667em\lower.7ex\hbox{E}\kern-.125emX}}

\usepackage{graphicx}       %for images
\usepackage{wrapfig}        %for images
\usepackage{makecell}       %for table
\usepackage{multirow}       %for table
\usepackage{booktabs}

\title{TCL: an ANN-to-SNN Conversion with Trainable Clipping Layers
}
\titleheader{To appear in the 58th Design Automation Conference (DAC 2021)}
\IEEEoverridecommandlockouts
\IEEEpubid{\makebox[\columnwidth]{978-1-6654-3274-0/21/\$31.00~\copyright2021 IEEE \hfill}\hspace{\columnsep}\makebox[\columnwidth]{ }}

\author{\IEEEauthorblockN{Nguyen-Dong Ho}
\IEEEauthorblockA{\textit{Kyung Hee University} \\
donghn@khu.ac.kr}
\and
\IEEEauthorblockN{Ik-Joon Chang}
\IEEEauthorblockA{\textit{Kyung Hee University} \\
ichang@khu.ac.kr}
}

\begin{document}
\maketitle
\IEEEpubidadjcol

\begin{abstract}
Spiking-neural-networks (SNNs) are promising at edge devices since the event-driven operations of SNNs provides significantly lower power compared to analog-neural-networks (ANNs). Although it is difficult to efficiently train SNNs, many techniques to convert trained ANNs to SNNs have been developed. However, after the conversion, a trade-off relation between accuracy and latency exists in SNNs, causing considerable latency in large size datasets such as ImageNet. We present a technique, named as TCL, to alleviate the trade-off problem, enabling the accuracy of 73.87\% (VGG-16) and 70.37\% (ResNet-34) for ImageNet with the moderate latency of 250 cycles in SNNs.
\end{abstract}


\section{Introduction}
\label{sec:introduction}
During the last decade, analog neural networks (ANNs) have shown rapid and extensive progresses. ANNs demonstrates their outstanding performance by surpassing the human-level accuracy for many applications such as image processing, voice recognition, and language translation. However, such ANN performance can be obtained at the cost of considerable power consumption. This makes it difficult to operate ANNs at resource-constraint edge devices. Unlike ANNs, spiking neural networks (SNNs) have event-driven behaviors, delivering significantly lower power dissipation. Consequently, researchers have considered SNNs as one of alternatives to ANNs for the resource-constraint edge devices. 

Nonetheless, the deployment of SNNs is limited since it is difficult to efficiently train SNNs. Due to the non-differential and discontinuous properties of SNNs, back-propagation cannot be applied for the training of SNNs. Some researchers have overcome this problem by using approximate techniques such as spike-base back-propagation \cite{huh_and_sej_2018, lee_2019, chankyu_lee_2020, yingyezhe_2018} and surrogate gradient \cite{wu_2018, bellec_2018, neftci_2019}. However, these techniques are only applicable to the training of small size networks for small size datasets. Further, when SNNs are trained based on the above techniques, forward and backward propagations need to be computed every time-step, unlike ANNs. As a result, the direct training approaches of SNNs suffer from considerably large overhead with respect to computational complexity and training time. 

Recently, some indirect training approaches of SNNs have been proposed, where the training results of ANNs are converted to SNN. For instance, Y. Cao et al. \cite{cao_2015} succeeded in converting ANNs to SNNs by mapping the output of rectified linear unit (ReLU) in ANNs to the spike rate in SNNs. Their technique shows good performances for the datasets of MNIST and CIFAR-10. P. U. Diehl et al. \cite{diehl_2015} developed the data-normalization technique that uses the model and the dataset of ANNs to estimate the normalization factors, leading to more improved conversion results. The authors of \cite{rueckauer_2017, sengupta_2019, rmp_snn_2020, snn_binary_2020} decided more accurate normalization factors by closely analyzing the relation between the activation of ANNs and the spike rate of SNNs. As a result, they successfully converted even large ANNs trained with the ImageNet dataset to SNNs. However, there is a trade-off relation between the accuracy and the latency of the converted SNNs, more problematic at the ImageNet dataset. As a result, the SNNs, converted by the above techniques, suffer from considerable accuracy degradation under low latency constraints. N. Rathi et al. \cite{nitinrathi_2020, dietsnn_2020} presented a novel technique, so called hybrid training, that retrains the SNNs obtained by ANN-to-SNN conversion, alleviating the trade-off relation. However, the addional SNN training phase of the hybrid training technique suffers from extremely heavy computations, as mentioned before, roughly 10 times larger compared to that of the ANN training \cite{nitinrathi_2020}.

Compared to \cite{cao_2015, diehl_2015, rueckauer_2017, sengupta_2019, rmp_snn_2020, snn_binary_2020, nitinrathi_2020, dietsnn_2020}, our work has significant contributions as follows. 
\begin{itemize}
    \item We formulate the reason why the trade-off relation between the accuracy and the latency of the converted SNNs exists, indicating how to mitigate the trade-off relation. 
    \item We propose a fundamental technique to ameliorate the trade-off relation between the accuracy and the latency of SNNs, trainable clipping layer (TCL), where the direct training of SNN is not required unlike \cite{nitinrathi_2020, dietsnn_2020}. We ensure that a clipping layer, whose clipping region is trained, follows a ReLU layer, finding the optimal data-normalization factor to consider both accuracy and latency in SNNs. 
    \item We further enhance the SNN accuracy under very low latency constraints by properly controlling a L2-regularization coefficient. 
\end{itemize}

From our experiments, SNNs based on our TCL technique show the following classification accuracies of CIFAR-10, 93.33\% at VGG-16 with 100-cycle latency and 92.06\% at ResNet-20 with 150 cycle latency. For the ImageNet dataset, the accuracies of VGG-16 and ResNet-34 are 73.87\% and 70.37\% respectively, with the latency of 250 cycles. At the very low latency below 40 cycles, VGG-16 provides the SNN accuracies of 92.60\% for CIFAR-10 and 70.75\% for ImageNet, well validating our proposed technique.  

\section{Preliminary backgrounds}
\label{sec:prelim}
\subsection{Spiking Neural Networks theory}
\label{subsec:snn-theory}
We can consider two representative SNN models, integrate-and-fire (IF), and leaky-integrate-and-fire (LIF) ones. It is well-known that the IF model is easily converted from an ANN, considered as the SNN model throughout our work. In the IF model, at time step $t$, the neuron $i$ in the $l^{th}$ layer has the summation of weighted spike input, $z^l_i$, as follows.        
\begin{equation}
\label{eq_sum_spike_input}
    z^l_i(t) =  \sum_{j} W^l_{ij} \theta^{l-1}_j(t) + b^l_i
\end{equation}
, where $W^l_{ij}$ is the synaptic weight, $b^l_i$ refers to the bias of the neuron, and $\theta^{l-1}_{j}$ is the spike input from the neuron $j$ that is in the previous layer. In the $l^{th}$ layer, the spike output of the neuron, $\theta^{l}_{i}$, remains zero until the membrane potential, $V^l_i$, reaches the threshold $V^l_{thr}$. At the time that $V^l_i$ becomes larger than or equal to $V^l_{thr}$, the spike output is fired. Hence, 
\begin{equation}
    \theta^l_i(t) =\begin{cases}
			1, & \text{if}\,\: V^l_i(t)\geq V^l_{thr}\, \\
            0, & \text{else.}
		 \end{cases}
\end{equation}
After the firing, the membrane potential, $V^l_i$, is reset. There are two approaches to reset $V^l_i$: reset-to-zero and reset-by-subtraction. Since the reset-to-zero suffers from considerable information loss \cite{rueckauer_2017}, the reset-by-subtraction is employed throughout this work. Under this situation, at the time step of $t$, $V^l_i$ can be modeled as follows.
\begin{equation}
\label{eq_neuron_reset}
    V^l_i(t)=V^l_i(t-1) + z^l_i(t) - V^l_{thr}\theta^{l}_{i}(t)
\end{equation}

\subsection{ANN to SNN conversion}
\label{subsec:snn-conversion}
The ReLU function is widely used as the activation function of ANNs, given by the following one.  
\begin{equation}
\label{eq_relu}
    a^l_i = max\left (0,\:\: \sum_{j} W^l_{ij}a^{l-1}_j + b^l_i\right )
\end{equation}
By comparing (\ref{eq_relu}) to (\ref{eq_sum_spike_input}), an ANN-to-SNN converting algorithm can be obtained. In SNNs, spike output signals are binary, only '1' or '0', implying that the spike outputs do not have negative values. Therefore, (\ref{eq_relu}) can be simply mapped to (\ref{eq_sum_spike_input}) by converting the ReLU output to the spike rates of SNNs. For the conversion, both weights and biases need to be normalized, namely data-normalization, where weights ($W^l$) and biases ($b^l$) are normalized by (\ref{eq_data_normalization}), and then, the threshold voltage of neurons is set to $1.0$.    
\begin{equation}
\label{eq_data_normalization}
    W^l = W^l\frac{\lambda^{l-1}}{\lambda^l} \text{\: and \:} b^l = \frac{b^l}{\lambda^l}
\end{equation}
 , where $\lambda^{l}$ is the normalization factor of the current layer, so called norm-factor, $\lambda^{l-1}$ is the norm-factor of the previous layer. The decision of norm-factors is more discussed in Section \ref{subsec:trade-off}. 

It is well-known that it is difficult to model max-pooling and batch-normalization in SNNs \cite{rueckauer_2017}. The max-pooling can be replaced with the other pooling techniques such as average-pooling, well-modeled in SNNs. The authors of \cite{rueckauer_2017} remove the batch-normalization, of (\ref{eq_batch_norm}) after the training of ANNs. 
\begin{equation}
\label{eq_batch_norm}
    BN(a)=\frac{\gamma}{\sigma}\left(a-\mu\right)+\beta
\end{equation}

, where $a$ is the input, $\mu$ and $\sigma$ are mean and variance of mini-batch, and $\gamma$ and $\beta$ are two learned parameters of batch-normalization. To prevent the accuracy loss of ANNs due to the removal of the batch-normalization, they scale weights and biases of corresponding convolution layers by using the following equation:
\begin{equation}
\label{eq_bn_remove}
    W^l_{ij} = \frac{\gamma^l_i}{\sigma^l_i}W^l_{ij} \text{\: and \:}
    b^l_{i} = \frac{\gamma^l_i}{\sigma^l_i}(b^l_{i} - \mu^l_{i}) + \beta^l_{i}.
\end{equation}

Our ANN-to-SNN conversion is based on the above discussion. We apply the data-normalization based on (\ref{eq_data_normalization}) and remove batch-normalization by using (\ref{eq_bn_remove}). We replace max-pooling layers with average-pooling ones. Instead of using a soft-max layer, not modeled in SNNs, we simply count the number of spiking signals and take the maximum to classify outputs. At the first SNN layer, we feed input signals with analog values, so called real coding, same as the technique which is used in \cite{rueckauer_2017, snn_binary_2020, dietsnn_2020}.

\section{Our contribution}
\label{sec:contribution}
In this section, we discuss how our TCL technique alleviates the trade-off relation latency and accuracy of SNNs. Before this, we explain the reason why the trade-off relation exists after ANN-to-SNN conversion.  

\subsection{The trade-off between accuracy and latency}
\label{subsec:trade-off}

\begin{figure}[htbp]
\centerline{\includegraphics[scale=0.64]{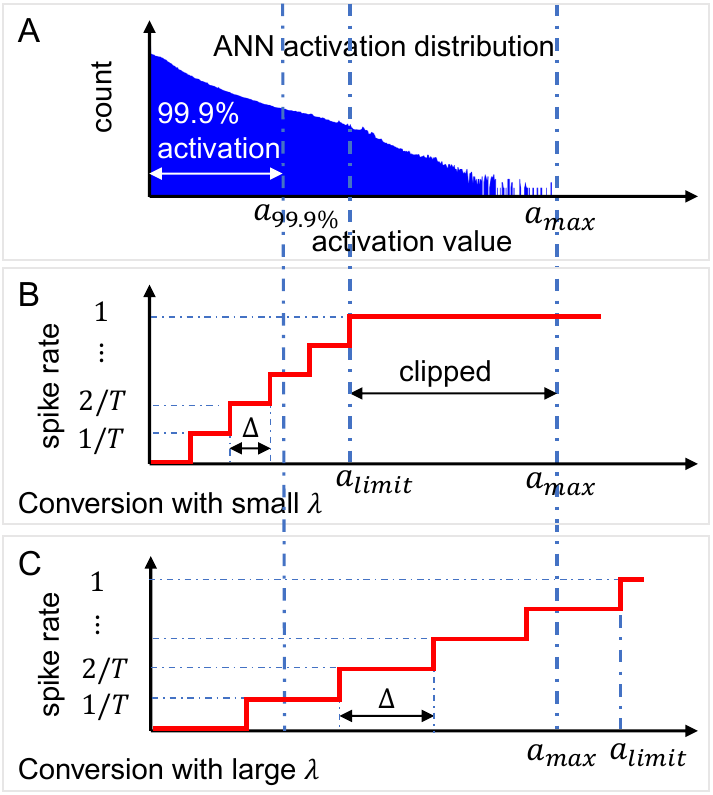}}
\caption{(A) Distribution of ANN activations (plotted in log-scale), (B) Conversion with small $\lambda$, (C) Conversion with large $\lambda$.}
\label{fig_mapping}
\end{figure}
In ANN-to-SNN conversion, an activation of ANNs, $a_i$, is mapped to a spike rate of SNNs, $f_i$, as mentioned above. The spike rate, $f_i$, can be estimated by following equation.

\begin{equation}
\label{eq_sprate_estimae}
    f_i \approx \begin{cases}
			\frac{ \left \lfloor \frac{a_i}{\Delta}  \right \rfloor}{T}, & \text{if}\,\: a_i\leq a_{limit}\,\: (\Delta = \frac{a_{limit}}{T})\\
            1.0, & \text{if}\,\: a_i> a_{limit}
		 \end{cases}
\end{equation}

 , where $a_{limit}$ is the maximum value of activations mapped to the maximum spike rate ($f_{max}=1$), and $\Delta$ is the quantization resolution. In ANN-to-SNN conversion, the activations of ANNs are quantized and then, the quantized activations are mapped to their corresponding spike rate, where some quantization error is unavoidable. To reduce the quantization error due to ANN-to-SNN conversion, we need to make smaller $\Delta$, which can be obtained by increasing the latency, $T$, or decreasing $a_{limit}$. This is the reason why after ANN-to-SNN conversion, latency and accuracy have trade-off relation. 
 In the ANN-to-SNN conversion with data-normalization, $\lambda$ is proportional to $a_{limit}$, where smaller $\lambda$ with the fixed $T$ tends to reduce $a_{limit}$. This results in the reduction of the quantization error. However, from a certain point, $a_{limit}$ is lower than $a_{max}$ and then, we have the conversion loss due to the clipping of activations, shown in Fig. \ref{fig_mapping}.B, degrading the accuracy of converted SNNs as well. On the other hand, larger $\lambda$ accompanies the increment of the quantization error, as shown in Fig. \ref{fig_mapping}.C.
 
The work in \cite{diehl_2015} determined the norm-factor of each layer by taking the maximum value among the activation parameters of the layer. However, in order to reduce quantization error, this approach requires large latency in SNNs. The authors of \cite{rueckauer_2017} and \cite{snn_binary_2020} decreased $\lambda$ by the following technique. In ANNs, most activations, roughly 99.0\% to 99.99\%, are placed in the range of of $[0,\; max/3]$. From this observation, they decide $\lambda$ by selecting the value of 99.9\%. The authors of \cite{sengupta_2019} and \cite{rmp_snn_2020} chose $\lambda$ based on the operation of SNN. They sequentially run SNN layers and take the maximum weighted spike output for $\lambda$, then scaled with the factor of $0.7\sim0.9$. Although the smaller $\lambda$ due to these techniques reduces quantization error, the conversion loss due to clipping error still causes significant accuracy degradation in SNNs.

In this work, we propose a novel technique to alleviate the trade off between quantization and clipping errors, where we decrease $a_{max}$ as small as possible while maintaining the accuracy of ANNs accuracy. Such an approach provides both low latency and high accuracy in SNNs, discussed in Section \ref{subsec:tcl}. 

\subsection{Trainable Clipping Layer}
\label{subsec:tcl}

\begin{figure}[htbp]
\centerline{\includegraphics[scale=0.62]{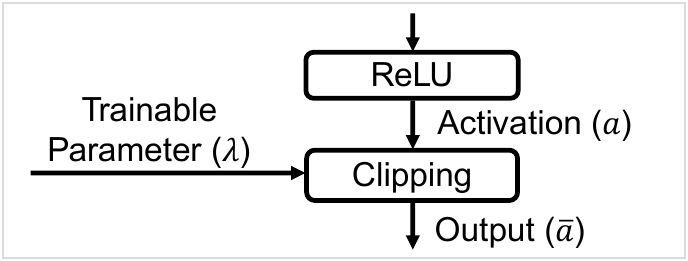}}
\caption{Clipping Layer for ANN activation}
\label{fig_clip_layer}
\end{figure}

%\begin{figure}[htbp]
%\includegraphics[scale=0.24]{act_histogram_small.pdf}
%\centering
%\caption{Distribution of ANN activations in the $2^{nd}$ layer of VGG-16 over the entire CIFAR-10 test-datasets. The distribution is plotted in log-scale.}
%\label{fig_acc_hist}
%\end{figure}

\begin{figure*}[!ht]
\includegraphics[scale=0.34]{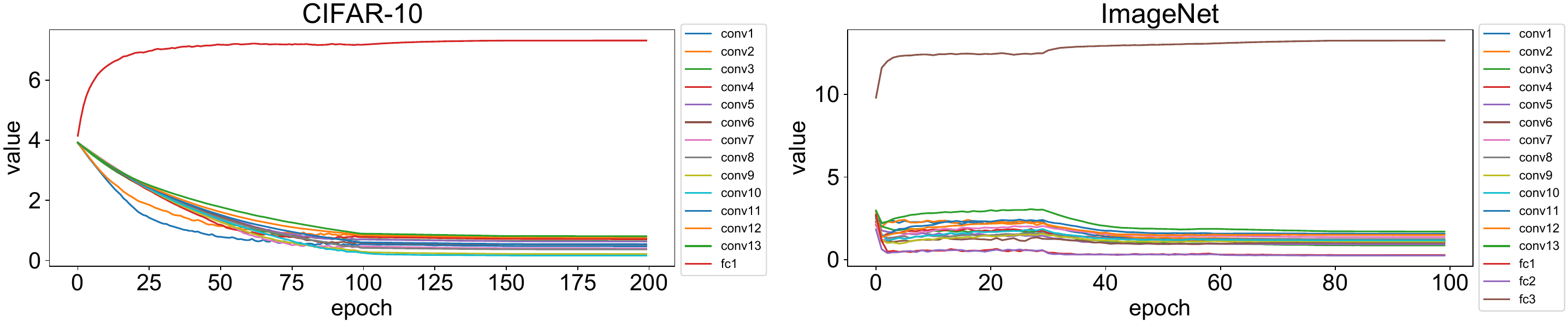}
\centering
\caption{Variation of each norm-factor during training VGG-16}
\label{fig_vgg16_norm_factor}
\end{figure*}

\begin{figure*}[!ht]
\includegraphics[scale=0.34]{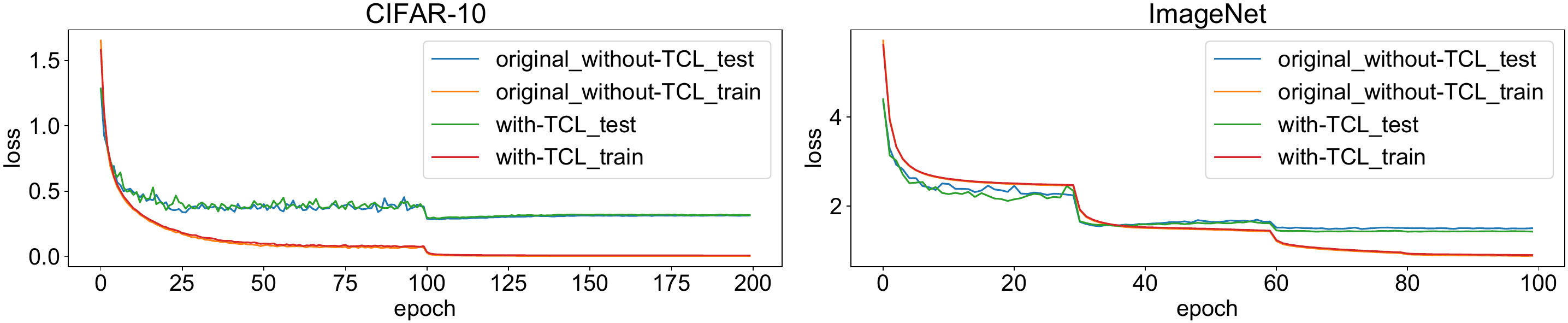}
\centering
\caption{Training and test losses during training VGG-16}
\label{fig_vgg16_loss}
\end{figure*}

\begin{figure*}[!ht]
\includegraphics[scale=0.34]{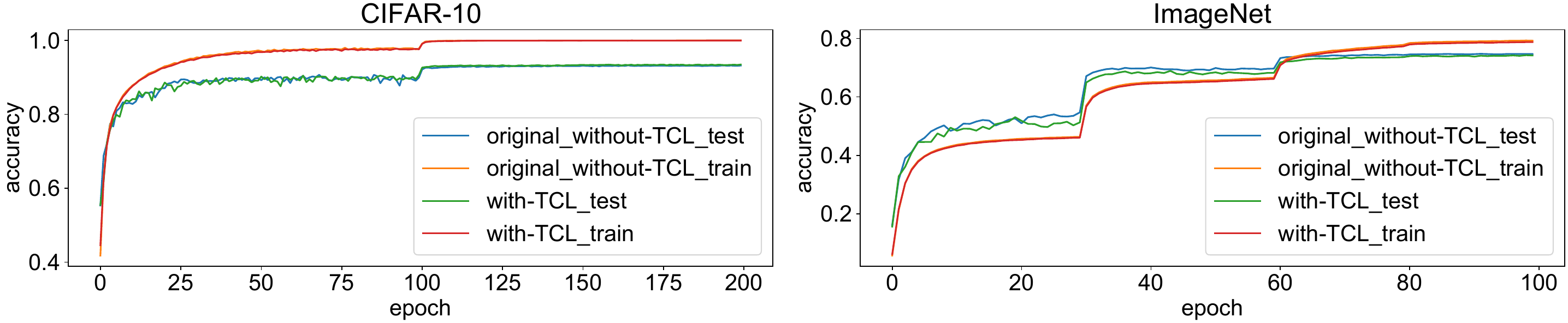}
\centering
\caption{Training and test accuracies during training VGG-16}
\label{fig_vgg16_acc}
\end{figure*}

To decide the norm-factor of a certain layer, instead of analyzing activations of the corresponding ANN layer, we add a clipping layer after the ReLUs of the ANN layer, shown in Fig. \ref{fig_clip_layer}. The forward function of the clipping layer is described in (\ref{eq_clip_forward}). Please, note that the clipping layer has a trainable parameter, $\lambda$, which becomes the norm-factor for the data-normalization. When backward computations of ANNs are processed, the gradients of $\lambda$ are formulated as (\ref{eq_clip_backward}). 
$\lambda$ is trained based on the learning rule of (\ref{eq_l2_norm}), where L2-regularization is used to optimize $\lambda$. 

\begin{equation}
\label{eq_clip_forward}
    \Bar{a} = clip(a, \,\lambda) = \begin{cases}
			\lambda, & \text{if } a\geq \lambda\, \\
            a, & \text{else.}
		 \end{cases}
\end{equation}

\begin{equation}
\label{eq_clip_backward}
	\frac{\partial \Bar{a}}{\partial\lambda} = \begin{cases}
			1, & \text{if } a\geq \lambda\, \\
            0, & \text{else.}
		 \end{cases}
\end{equation}

\begin{equation}
\label{eq_l2_norm}
\lambda\rightarrow\lambda-\eta\alpha\lambda-\eta\frac{\partial L}{\partial\lambda} = \lambda-\eta\alpha\lambda - \eta\frac{\partial\Bar{a}}{\partial\lambda}\frac{\partial L}{\partial\Bar{a}}
\end{equation}
 , where $\eta$ is the learning rate, $\alpha$ is the L2-regularization coefficient, and $L$ is the ANN loss. We visualize the training process of norm-factors, shown in Fig. \ref{fig_vgg16_norm_factor}. The results are obtained from the training of VGG-16 for two datasets, CIFAR-10 and ImageNet. In (\ref{eq_l2_norm}), the L2-regularization term ($-\eta\alpha\lambda$) tends to decrease $\lambda$, the norm-fact of the current layer, as already mentioned. From a certain point, the decrement of $\lambda$ causes training loss. Under this situation, the optimization term ($- \eta\frac{\partial\Bar{a}}{\partial\lambda}\frac{\partial L}{\partial\Bar{a}}$) increases $\lambda$ to complement the training loss. After some epochs, $\lambda$ starts to be stabilized to the optimal value with respect to the training loss. Fig. \ref{fig_vgg16_loss} shows training and test losses of each training epoch. We compare training and test losses of the original ANN training, where our TCL is not used, to those of our TCL-based ANN training. Further, we observe validation accuracies of each epoch, shown in Fig. \ref{fig_vgg16_acc}, clearly demonstrating that our TCL hardly affect the training results of VGG-16.

\begin{figure*}[!ht]
\includegraphics[scale=0.34]{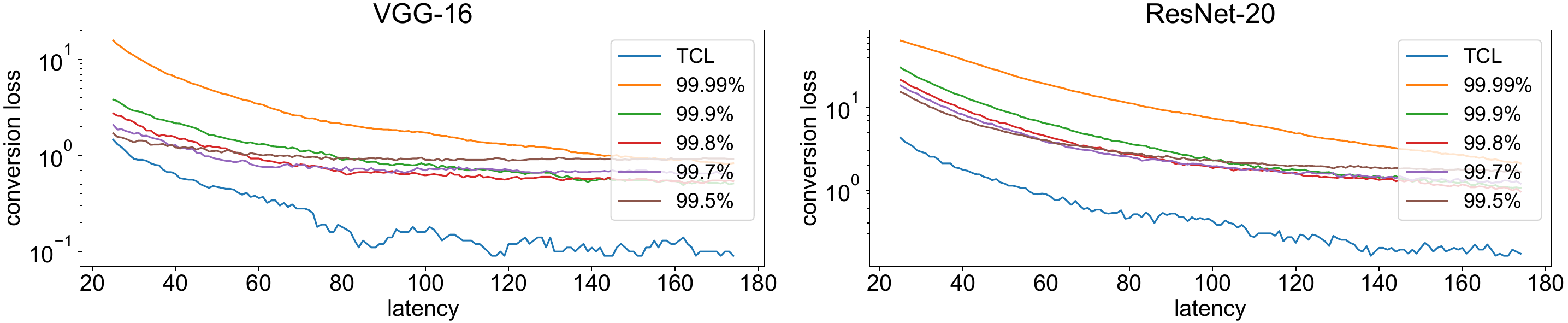}
\centering
\caption{Accuracy loss due to ANN-SNN conversion for the dataset of CIFAR-10}
\label{fig_conversion_loss}
\end{figure*}

To further prove the efficacy of our TCL, we investigate how our TCL affect the loss of ANN-to-SNN conversion by comparing the conversion loss due to our TCL to those due to various norm-factors, 99.99\%, 99.9\%, 99.8\%, 99.7\%, and 99.5\%, shown in Fig. \ref{fig_conversion_loss}. The conversion loss due to our TCL is significantly lower compared to the others, well showing that our TCL improves the accuracy of SNNs compared to \cite{rueckauer_2017, sengupta_2019, rmp_snn_2020, snn_binary_2020}.  

\section{Experimental results}
\label{sec:results}

We implemented our TCL on a Pytorch framework \cite{pytorch_framework}. We trained ANNs by using a stochastic gradient descent algorithm. We ensured that the total training epochs are 200 for CIFAR-10, and 100 for ImageNet. We initialize learning rates to $0.1$ besides VGG-16 and CNET with CIFAR-10, whose initial learning rates are $0.01$. We scaled the learning rates by $0.1$ at the training epochs of [100, 150] for CIFAR-10 and [30, 60, 80] for ImageNet. The initial value of $\lambda$ is set to $4.0$ for all networks with both CIFAR-10 and ImageNet datasets.

\subsection{The Effect of a L2-regularization Coefficient} 
\label{subsec:tcl-larger-alpha}
\begin{figure}[htbp]
\includegraphics[scale=0.32]{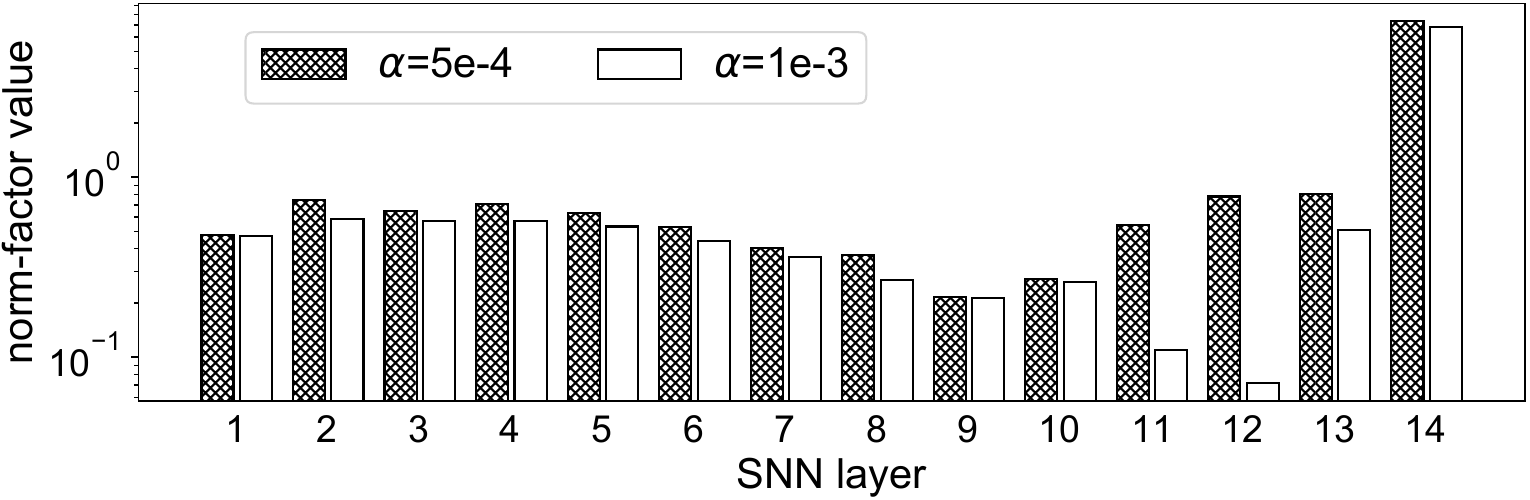}
\centering
\caption{The norm-factors of VGG-16 with two alpha values for CIFAR-10}
\label{fig_norm_with_different_alpha}
\end{figure}

In our TCL, the L2-regularization coefficient of (\ref{eq_l2_norm}), $\alpha$, affects the accuracy of SNNs with the following two styles. Firstly, as shown in Fig. \ref{fig_norm_with_different_alpha}, larger $\alpha$ tends to provide smaller norm-factors, resulting in enhanced SNN accuracy under very small latency constraints below 50 cycles. Secondly, in our TCL-based training the accuracy of ANNs shows the trend to become lower as $\alpha$ increases, potentially lowering corresponding SNN accuracies with moderate latency conditions above 200 cycles. The above two trends clearly appear in our experiment results, shown in Table \ref{tab_larger_alpha}. Since the above two trends oppositely influence the accuracy of SNNs, we need to find the optimal $\alpha$, the largest value not to affect the accuracy of ANNs. However, the algorithm to find the optimal values of hyper-parameters such as the L2-regularization coefficient is not well-established in previous researches yet, not explored in this work. From some trial and errors, we properly selected the values of $\alpha$ in this work. 

\begin{table}[!htbp]
  \caption{SNN accuracy at low latency}
  \centering
  \begin{tabular}{lcccccc}
    \midrule
    \multirow{2}{*}{Network}&\multirow{2}{*}{$\alpha$}&\multirow{2}{*}{\makecell{ANN\\accuracy}}&\multicolumn{4}{c}{SNN accuracy} \\
    \cmidrule(r){4-7}
    &&&T=20&T=25&T=30&T=35\\
    
    \midrule
    \multicolumn{6}{c}{\textbf{CIFAR-10}}\\
    \midrule
    VGG-16&5e-4&93.49&88.33&92.02&92.48&92.66 \\
    &1e-3&93.25&92.60&93.03&93.07&93.12 \\
    \cmidrule(r){1-7}
    ResNet-20&1e-4&92.20&85.55&87.88&89.30&89.88 \\
    &5e-4&91.58&89.63&90.50&90.98&91.22 \\
    \cmidrule(r){1-7}
    MobileNet&1e-4&91.68&-&-&10.02&20.53 \\
    &5e-4&91.17&-&-&64.29&87.50 \\
    
    \midrule
    \multicolumn{6}{c}{\textbf{ImageNet}}\\
    \midrule
    VGG-16&1e-4&73.91&-&57.42&64.20&67.59 \\
    &1e-3&73.22&-&65.72&69.36&70.75 \\
    
    \bottomrule
  \end{tabular}
  \label{tab_larger_alpha}
\end{table}

\subsection{Experiment results and discussion}
\label{subsec:results-discussion}
\begin{table*}[!htbp]
  \caption{Comparison of the results}
  \centering
  \begin{tabular}{llccccc}
    \midrule
    \multirow{2}{*}{Model}&\multirow{2}{*}{Architecture}&ANN accuracy&ANN accuracy&SNN&\multirow{2}{*}{Latency}&Conversion \\
    &&(without TCL)& (with TCL)&accuracy&&loss \\
    \midrule
    \multicolumn{7}{c}{\textbf{CIFAR-10}}\\
    \midrule
    
    Cao et al. (2015)\cite{cao_2015}&CNN (3Conv, 2Linear)&79.12&-&77.43&400&1.69\\
    \cmidrule(r){1-7}
    Rueckaur et al. (2017) \cite{rueckauer_2017}&CNN (4Conv, 2Linear)&91.91&-&90.85&400&1.06\\
    \cmidrule(r){1-7}
    Sengupta et al. (2019) \cite{sengupta_2019}&VGG-16&91.70&-&91.55&2500&0.15\\
    &ResNet-20&89.10&-&87.46&2500&1.64\\
    \cmidrule(r){1-7}
    Han et al. (2020) \cite{rmp_snn_2020}&VGG-16&93.63&-&93.51&1024&0.12\\
    &ResNet-20&91.47&-&91.36&2048&0.11\\
    \cmidrule(r){1-7}
    Rathi et al. (2020) \cite{nitinrathi_2020}&VGG-16&92.81&-&92.02&200&0.79\\
    &ResNet-20&93.15&-&92.22&250&0.93\\
    \cmidrule(r){1-7}
    Rathi \& Roy (2020) \cite{dietsnn_2020}&VGG-16&93.72&-&92.64&20&1.08\\
    &ResNet-20&92.79&-&92.14&25&0.65\\
    \cmidrule(r){1-7}
    \textbf{Ours}&CNET (4Conv, 2Linear)&89.96&90.05 ($\alpha=5\mathrm{e}{-4}$)&90.04&100&0.01\\
    \cmidrule(r){2-7}
    &VGG-16&93.28&93.49 ($\alpha=5\mathrm{e}{-4}$)&93.33&100&0.16\\
    &&&93.25 ($\alpha=1\mathrm{e}{-3}$)&92.60&20&0.65\\
    \cmidrule(r){2-7}
    &ResNet-20&92.26&92.20 ($\alpha=1\mathrm{e}{-4}$)&91.78&100&0.42\\
    &&&&92.06&150&0.14\\
    \cmidrule(r){4-7}
    &&&91.58 ($\alpha=5\mathrm{e}{-4}$)&90.50&25&1.08\\
    \cmidrule(r){2-7}
    &MobileNet&91.81&91.68 ($\alpha=1\mathrm{e}{-4}$)&91.26&150&0.42\\
    &&&&91.48&200&0.20\\
    \cmidrule(r){4-7}
    &&&91.17 ($\alpha=5\mathrm{e}{-4}$)&87.50&35&3.67\\

    \midrule
    \multicolumn{7}{c}{\textbf{ImageNet}}\\
    \midrule
    Rueckaur et al. (2017) \cite{rueckauer_2017}&VGG-16&63.89&-&49.61&400&14.28\\
    &Inception-V3&76.12&-&74.60&550&1.52\\
    \cmidrule(r){1-7}
    Sengupta et al. (2019) \cite{sengupta_2019}&VGG-16&70.52&-&69.96&2500&0.56\\
    &ResNet-34&70.69&-&65.47&2500&5.22\\
    \cmidrule(r){1-7}
    Lu \& Sengupta (2020) \cite{snn_binary_2020}&VGG-15&69.05&-&66.56&64&2.49\\
    \cmidrule(r){1-7}
    Han et al. (2020) \cite{rmp_snn_2020}&VGG-16&73.49&-&73.09&4096&0.40\\
    &ResNet-34&70.64&-&69.89&4096&0.75\\
    \cmidrule(r){1-7}
    Rathi et al. (2020) \cite{nitinrathi_2020}&VGG-16&69.35&-&65.19&250&4.16\\
    &ResNet-34&70.02&-&61.48&250&8.54\\
    \cmidrule(r){1-7}
    Rathi \& Roy (2020) \cite{dietsnn_2020}&VGG-16&70.08&-&66.52&25&3.56\\
    \cmidrule(r){1-7}
    \textbf{Ours}&VGG-16&73.85&73.91 ($\alpha=1\mathrm{e}{-4}$)&73.79&150&0.12\\
    &&&&73.87&250&0.04\\
    \cmidrule(r){4-7}
    &&&73.22 ($\alpha=1\mathrm{e}{-3}$)&70.75&35&2.47\\
    \cmidrule(r){2-7}
    &ResNet-34&70.87&70.85 ($\alpha=1\mathrm{e}{-4}$)&70.37&250&0.48\\
    &&&&70.66&300&0.19\\
    \cmidrule(r){2-7}
    &ResNet-50&75.42&75.33 ($\alpha=1\mathrm{e}{-4}$)&74.59&350&0.74\\
    \cmidrule(r){2-7}
    &MobileNet&70.54&66.82 ($\alpha=1\mathrm{e}{-4}$)&66.57&350&0.25\\
    
    \bottomrule
  \end{tabular}
  \label{tab_comparison}
\end{table*}

Our experiment results are summarized and compared to those of state-of-the-arts (SOTAs) related to ANN-to-SNN conversion in Table \ref{tab_comparison}. For CIFAR-10, \cite{cao_2015}, \cite{rueckauer_2017}, \cite{sengupta_2019}, and \cite{rmp_snn_2020} achieve good SNN accuracies after the ANN-to-SNN conversion. However, large latencies are required for those techniques. Although the authors of \cite{nitinrathi_2020, dietsnn_2020} alleviate the large latency problem by using their proposed hybrid training technique, where they reduce their latency to the cycles of 20$\sim$250, the accuracy loss due to the ANN-to-SNN conversion is still significant, larger than $0.5\%$. Our TCL technique makes the following two significant accomplishments compared to SOTAs.

\begin{itemize}
    \item In spite of limiting the range of activations, our TCL technique hardly affect the accuray of ANNs.
    \item After ANN-to-SNN conversion, SNNs show accuracies comparable to their ANN counterparts with moderate latency conditions. 
\end{itemize}
In the dataset of CIFAR-10, with the latency of 100$\sim$150 cycles, the accuracy loss due to the ANN-to-SNN conversion is almost negligible, less than $0.5\%$. The results of ImageNet further validate our contributions, where the training results of ANNs based on our TCL are almost same to their original accuracies. In addition, with the moderate latency of 250 cycles, we obtain good SNN accuracies, almost comparable to their ANN counterparts.

For very low latencies below 50 cycles , we train ANNs with a little larger $\alpha$. The results show almost comparable accuracy to the hybrid training of \cite{dietsnn_2020}, optimized for small latency constraints. Unlike \cite{dietsnn_2020}, where the direct SNN training phase in addition to the training one of ANN requires very large computational overhead, our TCL delivers good SNN accuracy with the only ANN training phase, our significant contribution as well.  

\section{Conclusion}
\label{sec:conclusion}
Many researches have shown that ANN-to-SNN conversion can become a realistic alternative to the direction training of SNNs. However, SNNs suffer from large latency, more problematic for large size dataset such as ImageNet, limiting the possibility of SNNs. In this work, we present a trainable clipping layer technique based on the ANN-to-SNN conversion, namely TCL, alleviating the trade-off relation between accuracy and latency of SNNs. Our experiment results shows that our TCL-based SNNs obtain almost comparable accuracy to their ANN counterpart for ImageNet even with the small latency of 250 clock cycles, well validating the efficacy of our TCL technique.

%\section*{Acknowledgment}
%This work was supported partly by Institute of Information \& communications Technology Planning \& Evaluation (IITP) grant funded by the Korea government(MSIT) (2020-0-01294, Development of IoT based edge computing ultra-low power artificial intelligent processor) and National Research Foundation of Korea (2020M3F3A2A01085755)

\bibliographystyle{unsrt}
\bibliography{refs}

\end{document}